\documentclass{article}
\usepackage{amsmath,graphicx,mlspconf}
\usepackage{xcolor}
\usepackage{adjustbox}
\usepackage{float}
\usepackage[colorinlistoftodos]{todonotes}
\usepackage{booktabs}
\usepackage{multirow}
\usepackage{amsmath}
\usepackage{amssymb}
\usepackage{hyperref}

\copyrightnotice{979-8-3503-2411-2/25/\$31.00 {\copyright}2025 IEEE}

\toappear{2025 IEEE INTERNATIONAL WORKSHOP ON MACHINE LEARNING FOR SIGNAL PROCESSING, Aug.\ 31-- Sep.\ 3, 2025, ISTANBUL, TURKEY}

\title{AbsoluteNet: A Deep Learning Neural Network to Classify Cerebral Hemodynamic Responses of Auditory Processing}

\name{Behtom Adeli*$^{1}$, John McLinden$^{1}$, Pankaj Pandey$^{1, 2}$, Ming Shao$^{3}$, Yalda Shahriari$^{1}$
\thanks{*Corresponding author: behtom@uri.edu\\ This work was supported by the National Science Foundation [NSF-2024418, NSF-2413573].}}

\address{\footnotesize$^{1}$Department of Electrical Engineering, Computer, \& Biomedical Engineering, University of Rhode Island RI, USA\\
\footnotesize$^{2}$Multimodal Functional Brain Imaging Research Lab at St. Jude Children's Research Hospital, TN, USA\\
\footnotesize$^{3}$Miner School of Computer and Information Sciences at University of massachusetts Lowell, MA, USA}

\begin{document}
\maketitle

\begin{abstract}
In recent years, deep learning (DL) approaches have demonstrated promising results in decoding hemodynamic responses captured by functional near-infrared spectroscopy (fNIRS), particularly in the context of brain–computer interface (BCI) applications. This work introduces AbsoluteNet, a novel deep learning architecture designed to classify auditory event-related responses recorded using fNIRS. The proposed network is built upon principles of spatio-temporal convolution and customized activation functions. Our model was compared against several models, namely fNIRSNET, MDNN, DeepConvNet, and ShallowConvNet. The results showed that AbsoluteNet outperforms existing models, reaching 87.0\% accuracy, 84.8\% sensitivity, and 89.2\% specificity in binary classification surpassing fNIRSNET, the second best model, by 3.8\% in accuracy. These findings underscore the effectiveness of our proposed deep learning model in decoding hemodynamic responses related to auditory processing and highlight the importance of spatio-temporal feature aggregation and customized activation functions to better fit fNIRS dynamics.
\end{abstract}


\section{Introduction}
\label{sec:intro}

In recent years, deep learning approaches (DL) – especially deep convolutional neural networks (CNNs) – have made significant strides in decoding hemodynamic responses captured by functional near-infrared spectroscopy (fNIRS) ~\cite{chen2022separation, sun2021dlfnirs, pandey2024fnirsnet}. CNNs can learn complex temporal features from preprocessed time series, obviating the need for manual feature extraction. For example, studies have shown that DL models such as EEGNet, MCNN, and Inception achieve promising results on neural data recorded using electroencephalography (EEG) \cite{2018eegnet,ma2021cnn,eastmond2022deep}. 

fNIRS measures hemodynamic responses via changes in the oxygenated (HbO$_2$) and deoxygenated (HbR) hemoglobin concentration, enabling noninvasive tracking of brain activity with a moderate trade-off in spatial and temporal resolution. This technique has been shown to reliably capture distinct hemodynamic patterns evoked by auditory stimuli ~\cite{mclinden2023individual, YooDecodingSound, ShatzerBrightening}, making it a valuable tool for studying auditory processing. In this contex, decoding and classifying auditory responses is crucial for characterizing and understanding sensory and cognitive processes~\cite{zhong2023classification}. Neural hemodynamic responses in the auditory cortex vary with different sound categories. Hemodynamic responses to stimuli, including auditory stimuli reflect the changing metabolic needs of active neuron populations, resulting in an influx of oxygenated hemoglobin followed by an undershoot and return to baseline ~\cite{mclinden2023phase}. While these responses are detectable, fNIRS data are often characterized by a low signal-to-noise ratio (SNR) and significant variability across trials and subjects. Physiological noise sources and motion artifacts can obscure the task-related signal~\cite{chen2022separation}. These challenges underscore the need for powerful computational techniques that can reliably discern single-trial auditory responses despite noise and inter-trial variability.

Due to the band-limited and sluggish nature of the hemodynamic response, architectures originally designed for EEG decoding—such as DeepConvNet and ShallowConvNet—do not achieve high performance unless they are adapted to the  spatial-temporal  characteristics of fNIRS data. For example, ShallowConvNet emphasizes low-frequency oscillatory power across channels, while DeepConvNet stacks multiple convolutional layers to learn hierarchical features from time series~\cite{schirrmeister2017deep}. These networks have somehow demonstrated physiologically meaningful feature extraction in fNIRS data as well, but they fail to achieve high accuracy rates.

Among fNIRS-specific models, CNN-based models have outperformed traditional classifiers in many fNIRS-based brain–computer interface (BCI) applications. For instance, a 3-class finger-tapping fNIRS-based BCI study achieved 92.7\% accuracy with a CNN, surpassing traditional support vector machine (SVM) and shallow neural network baselines on the same data~\cite{sun2021dlfnirs}. Another network called fNIRSNET utilized a dual-branch spatial-temporal CNN for classifying auditory hemodynamic responses~\cite{pandey2024fnirsnet}. In this network, one branch applies temporal followed by spatial convolution, while the other does the reverse. Their outputs were fused for final prediction, with HbO$_2$ and HbR concentration change signals serving as parallel input streams. Fully convolutional networks (FCNs) and Inception-based networks have also shown promise, using filters of varying lengths to capture multiscale neural dynamics recorded using fNIRS~\cite{ma2021cnn}. These models exploit the spatiotemporal characteristics of the fNIRS signals and improve generalization across subjects.

Building on these advances, we propose AbsoluteNet, a deep CNN model designed to enhance the classification of auditory processing recorded through fNIRS. This network is built on two streams of convolutional fusion blocks—one for processing spatial-temporal information and the other for temporal-spatial aggregation of information from fNIRS data. These are followed by concatenation and fusion blocks that further integrate the extracted features using separable convolution, average pooling, and a classification head. Finally, the network incorporates customized activation functions—namely, the squared, absolute, and logarithm of the absolute functions—each designed to have symmetric shapes around the y-axis\cite{adeli2025activation}. Leveraging the mentioned design implementations, AbsoluteNet aims to improve robustness and precision of single-trial neural decoding in fNIRS-based auditory processing classification task.

\begin{figure}[t]
  \centering
  \includegraphics[width=8.5cm]{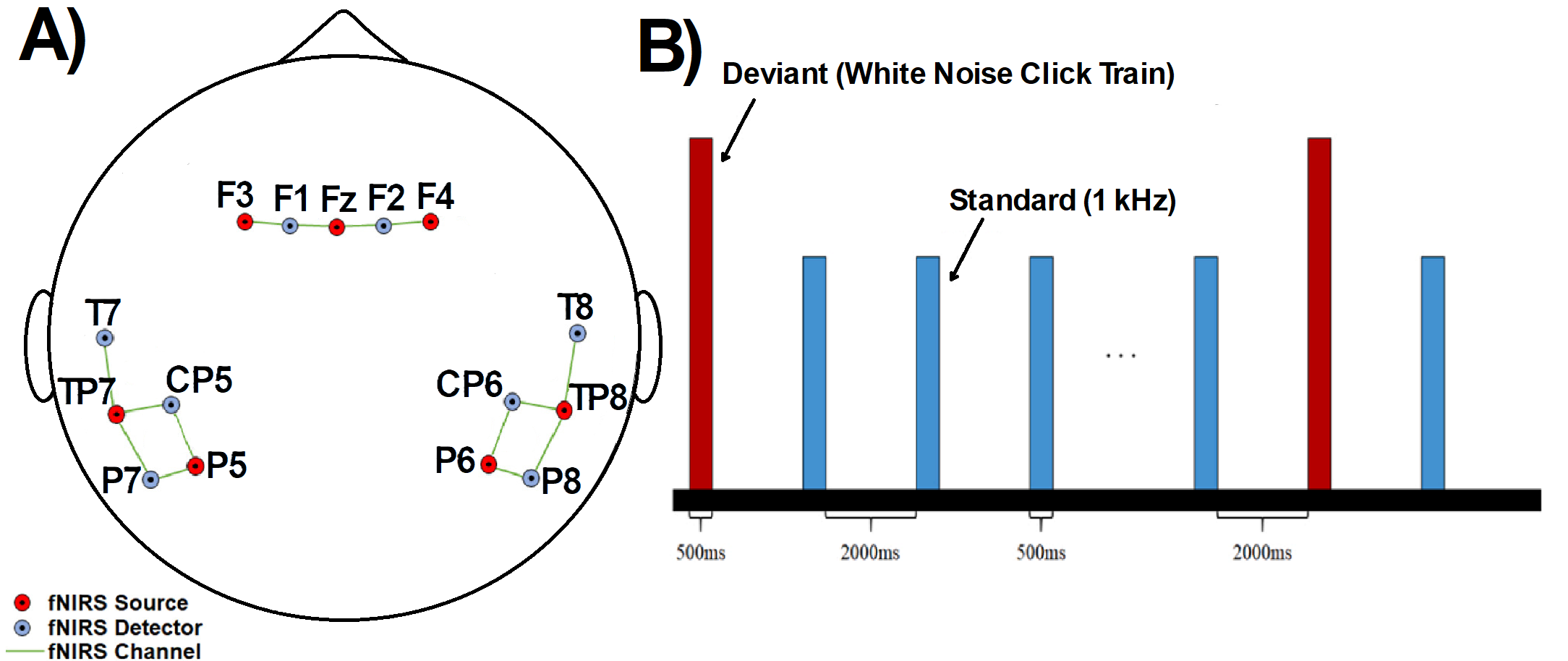}
  \caption{A) Source-detector placement and corresponding fNIRS channels. B) Auditory oddball paradigm with interleaved deviant (red) and standard (blue) tones and timing sequence.}
  \label{fig:data_recording}
\end{figure}

\section{Experimental Design and Data Preprocessing}
\label{sec:data}

fNIRS was used to measure hemodynamic responses during an auditory oddball task. Recordings were acquired using a NIRScout system (NIRx Inc.) with a sampling rate of $f_s = 7.8125$ Hz. Participants were seated comfortably in an armchair and instructed to attend a series of auditory stimuli. 14 fNIRS channels were recorded using seven sources and eight detectors, which were placed across three regions of interest (ROIs), including the frontal (F), left auditory (LA), and right auditory (RA) cortices as illustrated in Figure~\ref{fig:data_recording}A.

Recordings were obtained from nine healthy adults (four female, age: 28.33 $\pm$ 9.81 years), all with normal hearing and no history of neurological disorders. Each participant completed a single experimental session comprising six runs. An auditory oddball paradigm was employed, in which subjects were instructed to count the deviant tones embedded among standard stimuli. To further engage attention and semantic processing networks, subjects silently counted upward by five starting from each deviant tone, while imagining the enunciation of each number. Each run consisted of 20 deviant and 120--140 standard tones. To balance the dataset, standard trials occurring as the third or fourth tone between two deviants were randomly selected to match the number of deviant trials. Participants were asked to minimize movement and maintain visual fixation while breathing and blinking normally.

The auditory stimuli consisted of a 1 kHz standard tone and a 40 Hz white noise click train deviant tone, each lasting 500 ms, with a fixed inter-stimulus interval (ISI) of 2,000 ms (Figure ~\ref{fig:data_recording}B).

Raw light intensity signals were first converted to optical densities. To remove physiological noise, particularly cardiac artifacts near 1 Hz, a bandpass filter of 0.005--0.7 Hz was applied. Signals were then converted to HbO$_2$ and HbR concentration changes using the modified Beer–Lambert Law, as implemented in the nirsLab software package (NIRx Inc.) and upsampled to 10 Hz. The subsequent preprocessing including, trial rejection and epoching were performed using MATLAB software (MathWorks, Inc.). Each standard and deviant trial was segmented into a 15-second window following stimulus onset, resulting in epochs comprising 28 channels (14 for HbO$_2$ and 14 for HbR), with 150 temporal samples per trial (15 sec x 10 samples). This results in 1,080 trials  (9 subjects x 6 runs x 20 stimuli) per class (2 classes of (standard and deviant)). Noisy trials were rejected using visual inspection. A total of 918 clean trials per class were retained and pooled across all subjects for model training. 

\begin{figure*}[ht]
  \centering
  \includegraphics[width=\textwidth]{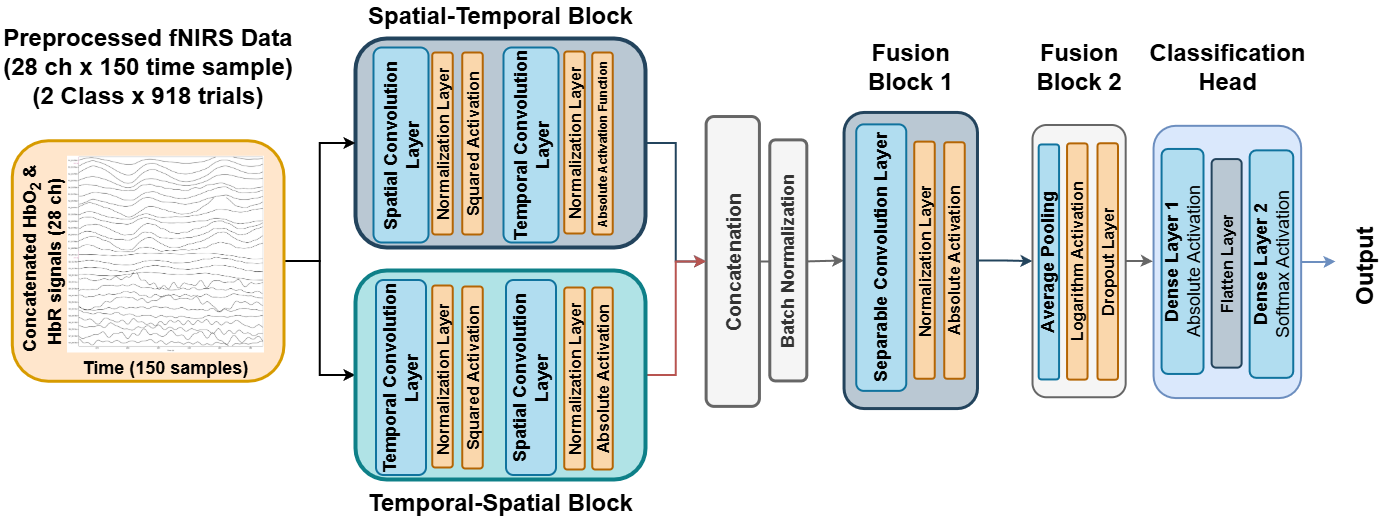}
  \caption{Block diagram of the AbsoluteNet architecture showing temporal-spatial dual streams, separable convolutions, and activation strategy.}
  \label{fig:absolutenet_arch}
\end{figure*}

\section{Our Proposed Deep Learning Model: AbsoluteNet}
\label{sec:method}
\subsection{Architecture}
The proposed deep neural network, AbsoluteNet, is based on Convolutional Neural Networks (CNNs) as shown in Figure~\ref{fig:absolutenet_arch}. The key network structures are as follows:\\
\textbf{Spatiotemporal Block Fusion:} Two combinations of temporal and spatial CNN kernels were employed and concatenated. The spatial-temporal block was used to aggregate information across all channels at each time sample before summarizing over the temporal dimension. Conversely, the temporal-spatial block was used to analyze temporal changes in each channel independently before integrating spatially across the scalp. Temporal convolution kernels were set to (1,5) and spatial convolution kernels to (28,1) corresponding to 500 ms and full-channel width, respectively. These kernel dimensions yielded superior accuracy compared to other configurations, as determined through hyperparameter optimization (see Section~\ref{subsec:training}). These parameters are consistent with prior works by Pandey et al.~\cite{pandey2024fnirsnet} and He et al.~\cite{he2022multimodal}.\\
\textbf{Fusion Block:} A separable convolutional layer was employed to efficiently extract and fuse high-level representations from the concatenated outputs of the spatial-temporal and temporal-spatial blocks. Unlike standard 2D convolution, which applies a single kernel across both spatial and temporal dimensions simultaneously, separable convolution factorizes the operation into two sequential steps—depth-wise and point-wise convolution—allowing for parameter efficiency and improved regularization.\\
In our formulation, each input sample is represented as a 3D tensor \( \mathbf{X} \in \mathbb{R}^{\text{ch} \times \text{t} \times C} \), where \(\text{ch} = 28\) denotes the number of fNIRS channels, \(\text{t} = 150\) is the number of time samples (15 sec x 10 samples), and \(C = 1\) is the input feature dimension (analogous to a grayscale channel in an image).\\A standard 2D convolution applies a kernel \\ \(\mathbf{K} \in \mathbb{R}^{k_{\text{ch}} \times k_{\text{t}} \times C_{\text{in}} \times C_{\text{out}}} \), where \(k_{\text{ch}}\), \(k_{\text{t}}\),
represent the convolutional window size which is the dimensions of the window along the spatial (channels) and temporal (time samples) dimensions that slides over the input during a convolution operation, and  \(C_{\text{in}}\), and \(C_{\text{out}}\) represent the number input and output feature layers of CNN (mentioned for each convolutional layer in table ~\ref{tab:absolutenet_layers}) respectively. This operation produces an output tensor \( \mathbf{Y} \in \mathbb{R}^{\text{ch}' \times \text{t}' \times C_{\text{out}}} \), computed as:
\begin{small}
\begin{equation}
\mathbf{Y}(ch,t,c_{\text{out}}) = 
\sum_{m=1}^{k_{\text{ch}}} \sum_{n=1}^{k_{\text{t}}} \sum_{c_{\text{in}}=1}^{C_{\text{in}}} 
\mathbf{X}(ch{+}m, t{+}n, c_{\text{in}}) \cdot \mathbf{K}(m,n,c_{\text{in}},c_{\text{out}})
\end{equation}
\end{small}

\noindent
where \(\mathbf{X}\) is the input to each convolutional layer, \(\mathbf{K}\) is the kernel size,  \( c_{\text{out}} \) indexes the output features, and \( (ch, t) \) denote the spatial (channels) and temporal (time samples) positions in the resulting feature map, respectively. The size of each kernel, and the inputs (output from the prior layer) are mentioned in Table ~\ref{tab:absolutenet_layers}. This structure allows better fusion of localized spatial-temporal features represented by the first two blocks. In contrast, separable convolution first performs a depth-wise convolution (one kernel per input feature layer) followed by a point-wise convolution (a \(1 \times 1 \) kernel to combine feature layers), significantly reducing the number of trainable parameters and computational complexity. This separation is particularly advantageous when combining the dual-branch features from spatial-temporal and temporal-spatial streams, as it avoids premature mixing of channel-specific dynamics while preserving distinct temporal and spatial characteristics before final integration.\\  
\textbf{Activation Functions:} Symmetrical activation functions over the y-axis were mainly used for our proposed model. Namely, the squared \( f(x) = x^2 \) and absolute \( f(x) = |x| \) were employed within the convolutional layers. The squared activation was applied in the initial convolutional blocks, followed by the absolute function in deeper layers. A logarithm of the absolute function \( f(x) = \log(|x|) \) was then applied after the average pooling layer of size (1,25) as optimized by genetic algorithm (GA) explained in section ~\ref{subsec:training}, enhancing feature compression and non-linearity in later stages of the network. This choice is supported by recent findings demonstrating its effectiveness on time series data \cite{alkhouly2021activation}.

\begin{table*}[ht]
\caption{\normalsize {Layer-Wise Architecture of the Proposed AbsoluteNet }}
\label{tab:absolutenet_layers}
\centering
\renewcommand{\arraystretch}{1.2}
\scriptsize
\begin{tabular}{lllllll}
\textbf{Block} & \textbf{Layer} & \textbf{Output Dimension} & \textbf{Kernel Size} & \textbf{Padding} & \textbf{Params} \\
\toprule

\multirow{1 }{*}{\textbf{Input}} & Concatenated HbO$_2$ + HbR & (28, 150, 1) & – & – & – \\

\midrule

\multirow{6}{*}{\textbf{Spatial-Temporal}} 
& Spatial Conv2D & (1, 150, 40) & (28, 1) & valid & 1,120 \\
& Layer Normalization & (1, 150, 40) & – & – & 80 \\
& Squared Activation & (1, 150, 40) & – & – & 0 \\
& Temporal Conv2D & (1, 146, 60) & (1, 5) & valid & 12,000 \\
& Layer Normalization & (1, 146, 60) & – & – & 120 \\
& Absolute Activation & (1, 146, 60) & – & – & 0 \\
\midrule

\multirow{6}{*}{\textbf{Temporal-Spatial}} 
& Temporal Conv2D & (28, 146, 20) & (1, 5) & valid & 100 \\
& Layer Normalization & (28, 146, 20) & – & – & 40 \\
& Squared Activation & (28, 146, 20) & – & – & 0 \\
& Spatial Conv2D & (1, 146, 60) & (28, 1) & valid & 33,600 \\
& Layer Normalization & (1, 146, 60) & – & – & 120 \\
& Absolute Activation & (1, 146, 60) & – & – & 0 \\
\midrule

\multirow{2}{*}{\textbf{Concatenation}} 
& Concatenation & (1, 146, 120) & – & – & 0 \\
& Batch Normalization & (1, 146, 120) & – & – & 480 \\
\midrule

\multirow{3}{*}{\textbf{Fusion Block 1}} 
& Separable Conv2D & (1, 146, 10) & (1, 3) & valid & 1,560 \\
& Layer Normalization & (1, 146, 10) & – & – & 20 \\
& Absolute Activation & (1, 146, 10) & – & – & 0 \\
\midrule

\multirow{3}{*}{\textbf{Fusion Block 2}} 
& Average Pooling 2D & (1, 16, 10) & (1, 25) & valid & 0 \\
& Logarithmic Activation & (1, 16, 10) & – & – & 0 \\
& Dropout (30\%) & (1, 16, 10) & – & – & 0 \\
\midrule

\multirow{3}{*}{\textbf{Classification Head}} 
& Dense (Absolute Activation) & (1, 16, 2) & – & – & 22 \\
& Flatten & (32,) & – & – & 0 \\
& Dense (Softmax) & (2,) & – & – & 66 \\
\midrule

\multicolumn{2}{r}{\textbf{Total Trainable Parameters}} & \multicolumn{2}{l}{\textbf{49,088}} & \multicolumn{2}{r}{\textbf{Total Parameters: 49,328}} \\

\bottomrule
\end{tabular}
\end{table*}

\subsection{Training Parameters and Optimization}
\label{subsec:training}
Python 3 and TensorFlow were used for the deep learning implementation. All training procedures were conducted on an NVIDIA GeForce RTX 4070 GPU.

\textbf{Training Parameters:} Cross-entropy loss and the Adam optimizer were used during training with an initial learning rate of  $9 \times 10^{-4}$ selected through hyperparameter optimization. The dropout rate was set to 0.3 to prevent the network from overfitting.

\textbf{5-Fold Cross Validation (CV):}  We deployed a 5-fold cross-validation (CV) method. The dataset was divided into 60\% for training, 20\% for validation, and 20\% for testing in each fold of the CV procedure. Each fold was trained for 200 epochs, and the model with the lowest validation loss was selected to be retrained for an additional 100 epochs using the combined training and validation data (80\%) and finally evaluated on the test set. The result is reported in terms of average with standard deviation across all the folds.

\textbf{Hyperparameter Optimization:} A genetic algorithm (GA) was employed to optimize the learning rate, convolutional kernel sizes, and average pooling dimensions. The GA was configured with a population size of 100, 30 generations, and a mutation rate of 0.1. An elitist selection strategy was used, in which the individuals with the lowest validation loss (i.e., highest fitness) were iteratively selected as parents for the next generation. Crossover was implemented by randomly recombining parent parameters, and mutation was applied by randomly perturbing individual genes within predefined bounds. The fitness was evaluated using the validation loss obtained over the first fold of a 5-fold cross-validation using the model trained from scratch for each individual.
\subsection{Ablation Study}
To further investigate the effects of the main blocks in our network we utilized four different ablation studies. In the $1^{\text{st}}$ and $2^{\text{nd}}$ studies we removed the temporal-spatial and spatial-temporal convolutional blocks, respectively. In the $3^{\text{nd}}$ study we removed the first fusion block containing the separable convolutional layer as well as the subsequent normalization and activation. Lastly, we removed the second fusion block comprised of average pooling, the logarithm of absolute activation function, and the dropout.
\section{Results}
\label{sec:results}

\begin{figure}[t]
  \centering
  \includegraphics[width=8.5cm]{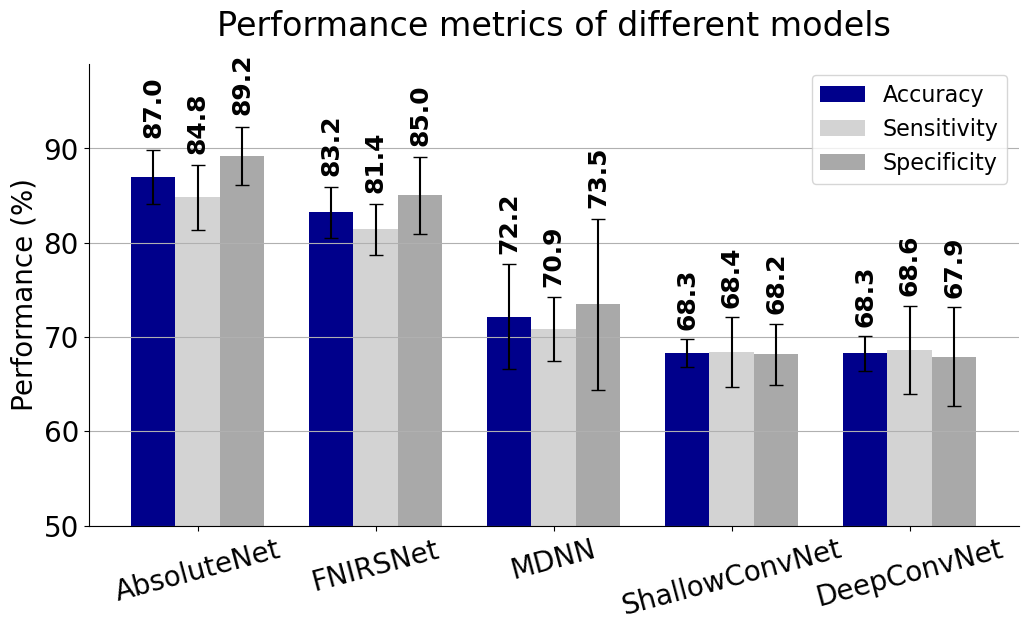}
  \caption{Performance comparison across multiple deep learning models on the fNIRS dataset.}
  \label{fig:output}
\end{figure}

Table~\ref{tab:performance_metrics} summarizes the classification metrics for AbsoluteNet using different input configurations. As shown, AbsoluteNet achieved the highest performance using concatenated HbO$_2$ and HbR.  Figure ~\ref{fig:output} demonstrates the average accuracy, sensitivity, and specificity results for the tested networks. All networks were re-calibrated to fit the dataset and the concatenated HbO$_2$ and HbR dataset \(28 \times 150\) was used as the input for all networks. As shown, AbsoluteNet demonstrated  a notable improvement in accuracy (87.0\%), sensitivity (84.81\%), and specificity (89.21\%),  surpassing fNIRSNET by 3.8\%, 3.4\%, and 4.2\%, respectively. These gains are particularly meaningful in the context of fNIRS-based signal processing and classification, where low SNRs and inter-subject variability pose persistent challenges. The heightened sensitivity reflects AbsoluteNet’s robustness in correctly detecting deviant stimuli, while the improved specificity indicates its strength in minimizing false positives. Together, these metrics underscore AbsoluteNet's balanced and practical performance for neural decoding applications.

\begin{table}[ht]
\caption{Performance Metrics Across Input Types}
\label{tab:performance_metrics}
\centering
\footnotesize
\resizebox{\columnwidth}{!}{%
\begin{tabular}{lccc}
\toprule
\textbf{Input} &  \textbf{Accuracy} (\%)& \textbf{Sensitivity} (\%) & \textbf{Specificity}  (\%)\\
\midrule
HbO$_2$       & 82.73 $\pm$ 3.34  & 81.21 $\pm$ 3.72  & 84.26 $\pm$ 3.95 \\
HbR       & 83.69 $\pm$ 4.89  & 80.66 $\pm$ 4.27  & 86.70 $\pm$ 5.95 \\
\textbf{HbO$_2$+HbR}   & \textbf{87.50 $\pm$ 2.88}  & \textbf{84.81 $\pm$ 3.45}  & \textbf{89.21 $\pm$ 3.10} \\
\bottomrule
\end{tabular}
}
\end{table}

\begin{figure}[t]
  \centering
  \includegraphics[width=8.5cm]{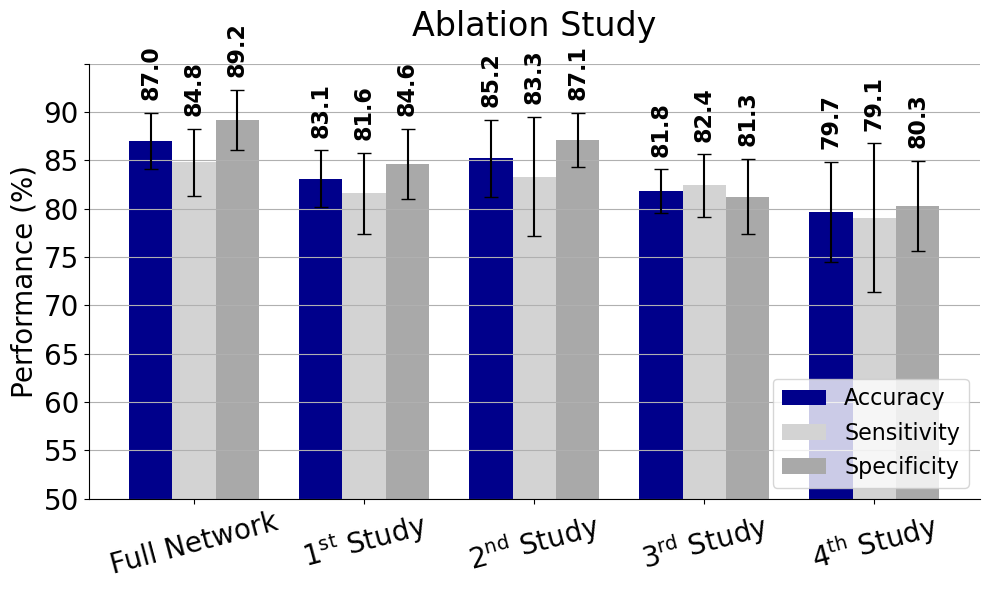}
  \caption{Ablation studies' performance evaluating the impact of removing key components in AbsoluteNet.}
  \label{fig:ablation}
\end{figure}

In experiments with single-modality inputs—HbO$_2$ and HbR—the network maintained relatively strong performance, with HbR yielding slightly better specificity than HbO$_2$. To accommodate single-channel input, the spatial kernel size was reduced from 28 (for HbO$_2$+HbR) to 14, ensuring the convolutional operation remained spatially valid. The concatenated dual-input configuration (HbO$_2$+HbR), however, achieved the best results overall, suggesting that joint representation of both chromophores improves feature richness and enhances classification accuracy. This reinforces the value of integrating complementary hemodynamic signals in fNIRS-based neural models. Ablation analysis (Figure~\ref{fig:ablation}) highlights the contribution of individual components within the AbsoluteNet architecture. Removing the first temporal-spatial block ($1^{\text{st}}$ Study) resulted in a reduction of accuracy to 83.1\%, indicating its importance in capturing temporal dynamics before spatial integration. Similarly, removing the spatial-temporal branch ($2^{\text{nd}}$ Study) reduced accuracy to 85.2\%, suggesting both dual pathways contribute complementarily to performance with more drop in accuracy if the temporal-spatial block is dropped.

The most severe degradation occurred when the second fusion block (i.e., the average pooling, logarithmic activation function coupled with the dropout rate) was removed ($4^{\text{th}}$ Study), leading to substantial  drop in accuracy (79.7\%), sensitivity (79.1\%), and specificity (80.3\%). This was followed by the results produced by the third study, the removal of the first fusion block, comprised of the separable convolutional layer, normalization layer, and the absolute activation function with accuracy drop of 5.2 \%, 2.4\%, and 7.9\% in accuracy, sensitivity, and specificity respectively. This demonstrates the critical role of the fusion blocks in combining and refining features post-concatenation.

\section{Discussion and Conclusion}
\label{sec:discussion}

This study introduces AbsoluteNet, a deep neural network tailored to extract features using spatial-temporal and temporal-spatial combination blocks with customized activation functions, symmetrical to the y-axis, from fNIRS recordings of auditory stimuli. The proposed model was specifically designed to include separate convolutional blocks to elucidate temporal-spatial and spatial-temporal features in combination with the use of absolute, square, and logarithmic activation functions proved essential for capturing the distinctive features of the fNIRS data, showcased by improved accuracy and class separability over baseline models such as fNIRSNET, ShallowCNN, DeepConvNet. These activation functions provided greater non-linearity, by leveraging the negative results of the convolutional layers, thus enhancing the model's sensitivity to subtle signal variations in low-SNR fNIRS data.

The proposed network demonstrated its ability to classify human auditory responses to standard vs. deviant tones from single-trial recordings, even with a limited number of training samples. Deep convolutional layers and temporal-spatial convolution design, coupled with separable layers, allowed the model to extract rich hemodynamic features across channels and time. The inclusion of GA-based hyperparameter tuning contributed further to performance gains. As shown in ~\ref{tab:performance_metrics}, the performance metrics peaks when the network is fed both HbO$_2$ and HbR signals. This might indicate that each signal has differentiable signatures that can be added up to reach a higher accuracy. Overall, AbsoluteNet showed capabilities in setting a new benchmark for single-modality fNIRS classification in auditory paradigms and provided a promising foundation for practical fNIRS-based data classifications.

This study faces a few limitations. First, the relatively small dataset constrains the generalizability of our findings in decoding auditory processing using fNIRS. Future work will address this by evaluating the proposed deep learning framework on larger and more diverse datasets to improve model robustness and performance across subjects. Second, the current analysis relies solely on fNIRS data, which provides limited temporal resolution. Future research should explore early fusion strategies that integrate complementary modalities, such as EEG, to leverage both the fast temporal resolution of EEG and the spatial specificity of fNIRS. This multimodal approach has the potential to enhance classification accuracy and offer a more comprehensive understanding of auditory processing dynamics. An additional limitation arises from the experimental design, which could benefit from a block design so that the standard and deviant trials would have 15 seconds of complete response followed possibly by a 15 seconds rest before the start of the next trial.

\bibliographystyle{IEEEbib}
\bibliography{refs}

\end{document}